# Structured Learning from Partial Annotations

**Xinghua Lou**  XINGHUA.LOU@IWR.UNI-HEIDELBERG.DE
**Fred A. Hamprecht**  FRED.HAMPRECHT@IWR.UNI-HEIDELBERG.DE
HCI, IWR, University of Heidelberg, Heidelberg 69115, Germany

## Abstract

Structured learning is appropriate when predicting structured outputs such as trees, graphs, or sequences. Most prior work requires the training set to consist of *complete* trees, graphs or sequences. Specifying such detailed ground truth can be tedious or infeasible for large outputs. Our main contribution is a large margin formulation that makes structured learning from only *partially* annotated data possible. The resulting optimization problem is non-convex, yet can be efficiently solve by concave-convex procedure (CCCP) with novel speedup strategies. We apply our method to a challenging tracking-by-assignment problem of a variable number of divisible objects. On this benchmark, using only 25% of a full annotation we achieve a performance comparable to a model learned with a full annotation. Finally, we offer a unifying perspective of previous work using the hinge, ramp, or max loss for structured learning, followed by an empirical comparison on their practical performance.

## 1. Introduction

Given a training set, structured learning extracts rules that allow the prediction of complex, structured output from structured input. It has improved conceptual clarity of, and boosted performance in, different tasks such as image segmentation, graph matching, word alignment, grammatical tagging, protein structure prediction or cell tracking (see (Bakir et al., 2006) and references therein). However, most classic structured learning algorithms require a strong prerequisite: the training data with complex structure needs to be **completely** annotated in order to apply those algorithms. This results in expensive, time-consuming data preparation and makes re-training very difficult.

In this paper, we study the problem of structured learning from **partial** annotations. Our contributions are the proposal of the new "bridge" loss function (section 3.2), a synthesis of previous work (section 3.4), a large margin problem formulation (section 3.5) with a suitable optimization strategy (section 4) and experimental validation (section 5).

### 1.1. Prior Art

We build on important previous work for multiclass classification with ambiguous labels. Here, each training sample $x$ has an exact, unknown label $c^*$. However, the training set only comprises a set of candidate labels $\boldsymbol{c}^*$ for each observation, where $c^* \in \boldsymbol{c}^*$. (Jin & Ghahramani, 2002) proposed an EM-like algorithm that iteratively estimates the label distribution and classifies using this distribution as a prior. Recently, (Cour et al., 2011) proposed convex loss for partial labels, which in turn resembles the one-versus-all loss (Zhang, 2004). We will extend this loss to structured data and discuss its properties in section 3.2.

This work is also closely related (section 3.3) to structured learning with latent variables (Yu & Joachims, 2009; Girshick et al., 2011). The main differences with (Yu & Joachims, 2009) are, theoretically, the derivation as an extension of multiclass learning with ambiguous labels and, practically, an improved optimization strategy (section 5). Structured learning for partially annotated sequences is studied in (Fernandes & Brefeld, 2011). These authors use a structured perceptron which, according to the experiments in section 5, does not generalize as well as the proposed large-margin method.

Finally, note that structured learning from partial annotations is different from semi-supervised or unsupervised structured learning (Altun et al., 2006; Xu et al., 2006; Zien et al., 2007). In those settings, training samples are either completely annotated or comple-ly unannotated.

## 2. Structured Learning from Partial Annotations

We want to learn from a *partially* annotated training set $\{(x_n, y_n^*) \in \mathcal{X}_n \times \mathcal{Y}_n : n = 1, \ldots, N\}$. Here, $x_n$ is a structured input from a space $\mathcal{X}_n$[1]. $y^*$ is a *partially* annotated structured output which induces a partitioning of structured output space $\mathcal{Y}$ into two sets $\mathcal{Y}^* \cap \mathcal{Y}^\circ = \emptyset$, $\mathcal{Y}^* \cup \mathcal{Y}^\circ = \mathcal{Y}$. $\mathcal{Y}^*$ comprises all outputs that *are* compatible with a partial annotation $y^*$, while $\mathcal{Y}^\circ$ encompasses all those structured outputs that are *not* compatible with the partial annotation, see Fig. 1.

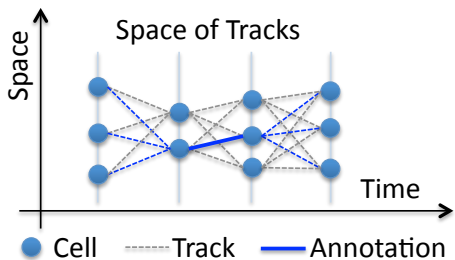

*Figure 1.* Tracking cell (circles) by assignment. Any path from left to right is a possible track. They form the complete track space $\mathcal{Y}$. Partial annotation (blue solid thick) divides this space into paths passing this annotation (any path in blue), i.e. $\mathcal{Y}^*$, and those not passing it (any path in gray), i.e. $\mathcal{Y}^\circ$. (best view in color)

## 3. Large Margin Learning from Partial Annotations

The aim of learning is to find a parameter vector $w$ that minimizes the weighted sum of a regularization term $\Omega(w)$ and the empirical loss:

$$\min_{w} \ J(w) := \lambda \Omega(w) + \frac{1}{N} \sum_{n=1}^{N} L(x_n, y_n^*; w), \quad (1)$$

The particular choice of regularizer and loss function leads to different learning methods, with the squared norm and hinge loss, respectively, yielding a structSVM (Tsochantaridis et al., 2006), which is particularly popular in structured output learning.

In the following, we propose two formulations of large margin learning from partial annotations. The first is a formal generalization of one-versus-all learning to structured data which is instructive, but not viable in practice. The second is a tractable formulation which is a generalization of multiclass learning from ambiguous labels. Hurried readers may want to jump to this proposal in section 3.2.

---

[1] Note that the cardinality of the spaces $\mathcal{X}_n$, $\mathcal{Y}_n$ is typically different for each input $n$.

*Table 1.* Summary of notation.

| Symbol | Definition |
|---|---|
| $x$ | Input data, structured or flat |
| $l(a)$ | $\|1-a\|_+$, hinge loss function |
| $c$ | A flat output (class label) |
| $c^*$ | True class label in multiclass learning |
| $\boldsymbol{c}^*$ | Set of ambiguous class labels, $c^* \in \boldsymbol{c}^*$ |
| $\boldsymbol{C}$ | Set of all class labels, $\boldsymbol{c}^* \subseteq \boldsymbol{C}$ |
| $y$ | A structured output |
| $y^*$ | Partially annotated structured output ground truth |
| $\mathcal{Y}$ | Space of all structured outputs |
| $\mathcal{Y}^*$ | Subspace induced by, and consistent with, partial annotation $y^*$ |
| $\mathcal{Y}^\circ$ | $\mathcal{Y}^\circ = \mathcal{Y} \setminus \mathcal{Y}^*$ Subspace of all structured outputs that are *not* compatible with $y^*$ |
| $\Delta(y^*, y)$ | Task loss function, measures the discrepancy between some $y$ and partial ground truth $y^*$ |
| $f(x, y; w)$ | Score of a prediction tuple $(x, y)$ given learned parameter vector $w$ |

### 3.1. Formulation I: One-Versus-All

The recently proposed "convex loss for partial labels" (CLPL) (Cour et al., 2011) enjoys favorable properties, including convexity, consistency and demonstrably high performance for flat (unstructured) outputs. Using notations in Table 1, CLPL can be expressed as

$$L^{\text{clpl}}(x, \boldsymbol{c}^*; w) = l\left(\frac{1}{|\boldsymbol{c}^*|} \sum_{c \in \boldsymbol{c}^*} f(x, c; w)\right) + \sum_{c \in \boldsymbol{C} \setminus \boldsymbol{c}^*} l\left(-f(x, c; w)\right) \quad (2)$$

Generalizing this loss to structured outputs gives

$$L^{\text{clpl-sl}}(x, y^*; w) = l\left(\frac{1}{|\mathcal{Y}^*|} \sum_{y \in \mathcal{Y}^*} f(x, y; w)\right) + \sum_{y \in \mathcal{Y}^\circ} l\left(-f(x, y; w)\right) \quad (3)$$

where $l$ is the hinge loss function and $f(x, y; w)$ the score of a prediction $y$ given input $x$ and parameter vector $w$.

Inserting this loss into Eq. 1 and using slack variables to prevent overfitting as in (Tsochantaridis et al.,

2006), we obtain the following optimization problem:

$$\min_{\boldsymbol{w}} \lambda\Omega(\boldsymbol{w}) + \frac{1}{N}\sum_n \xi_n + \frac{1}{N}\sum_n \sum_{y_n^\circ \in \boldsymbol{\mathcal{Y}}_n^\circ} \xi_{n,y_n^\circ}$$
$$\begin{aligned}
\text{s.t. } \forall n, \quad & \mathbb{E}_{y|y\in\boldsymbol{\mathcal{Y}}_n^*}[f(x_n,y;w)] \geq 1-\xi_n \\
\forall n, \forall y_n^\circ \in \boldsymbol{\mathcal{Y}}_n^\circ, \quad & -f(x_n,y;w) \geq 1-\xi_{n,y_n^\circ} \\
\forall n, \quad & \xi_n \geq 0 \\
\forall y_n^\circ \in \boldsymbol{\mathcal{Y}}_n^\circ, \quad & \xi_{n,y_n^\circ} \geq 0
\end{aligned}$$

The above formulation is convex, yet intractable in practice. Firstly, there are exponentially many terms in the target function, which thus cannot be represented explicitly. Secondly, and more importantly, this formulation requires computing the conditional expectations of the scores over the entire spaces $\boldsymbol{\mathcal{Y}}_n^*$ which, similar to estimating the partition function of graphical models, is usually intractable. To circumvent this limitation, one could use pseudo-likelihood (maximizing the margin is similar to maximizing the likelihood ratio) to approximate this expectation as in (Lee et al., 2006). However, a recent study shows that this is only possible when the distribution of training sample is rich enough (Sontag et al., 2010). This assumption is in contradiction to our problem setting, where annotations are partial and rare.

### 3.2. Formulation II: Pairwise Comparison

When formulated as a conventional multiclass learning problem, structured learning needs to discriminate a correct structured output from an exponential number of wrong structured outputs. As a way out, structSVM (Tsochantaridis et al., 2006) penalizes small or negative margins (differences) between the score of the correct structured output and the highest score among any of the wrong structured outputs.

We follow the same argument, by constructing a loss function that penalizes small margins between the current prediction (maximizer of the first term in the loss below) and the best scoring wrong prediction (maximizer of the second term in the following loss function):

$$L^{\text{pair}}(x,y^*;\boldsymbol{w}) = l\left(\max_{y\in\boldsymbol{\mathcal{Y}}^*} f(x,y;\boldsymbol{w}) - \max_{y\in\boldsymbol{\mathcal{Y}}^\circ} f(x,y;\boldsymbol{w})\right)$$

Note that, unlike the CLPL in Eq. 3, this loss function is not convex. Even so, we show in section 5 that the resulting learning problem can still be solved efficiently with an improved concave-convex procedure.

(Tsochantaridis et al., 2006) suggest to adjust the penalty levied for high-scoring wrong predictions according to just how wrong they are, as measured by a user-defined task loss function $\Delta(y^*,y)$. The idea is to push the decision boundary away from "bad" predictions. Following this idea, we obtain

$$\begin{aligned}
L^{\text{bridge}}(x,y^*;\boldsymbol{w}) &= \Big| \max_{y\in\boldsymbol{\mathcal{Y}}^\circ}[f(x,y;\boldsymbol{w}) + \Delta(y^*,y)] \\
&\quad - \max_{y\in\boldsymbol{\mathcal{Y}}^*}[f(x,y;\boldsymbol{w})] \Big|_+ \quad (4)
\end{aligned}$$

which we name the **bridge loss**, given that its margin is always computed across two disjoint spaces.

### 3.3. Connection to structSVMs with Latent Variables

Even though the loss in Eq. 4 was motivated and derived from multiclass learning with ambiguous labels, it has a very similar structure to the loss defined in the context of structured learning with latent variables. Specifically, (Yu & Joachims, 2009) address the problem when a training sample $(x,z^*,h) \in \boldsymbol{\mathcal{X}} \times \boldsymbol{\mathcal{Z}} \times \boldsymbol{\mathcal{H}}$ consists of both observed variables $z^* \in \boldsymbol{\mathcal{Z}}$ and unknown hidden variables $h \in \boldsymbol{\mathcal{H}}$. They propose a hinge loss with latent variables:

$$\begin{aligned}
L^{\text{hinge}}(x,z^*;\boldsymbol{w}) &= \max_{(z,h)\in\boldsymbol{\mathcal{Z}}\times\boldsymbol{\mathcal{H}}}[f(x,z,h;\boldsymbol{w}) + \Delta(z^*,z)] \\
&\quad - \max_{h\in\boldsymbol{\mathcal{H}}}[f(x,z^*,h;\boldsymbol{w})] \quad (5)
\end{aligned}$$

Indeed, in our problem setting the un-annotated part of a structure can be considered as a collection of hidden variables. In that sense, $\boldsymbol{\mathcal{Z}} \times \boldsymbol{\mathcal{H}}$ corresponds to $\boldsymbol{\mathcal{Y}}$ in our setting and, by fixing the observed variables, $z^*$, $z^* \times \boldsymbol{\mathcal{H}}$ amounts to $\boldsymbol{\mathcal{Y}}^*$. Now the key difference between Eq. 4 and Eq. 5 is clear: the bridge loss searches the most misleading (high-scoring, but wrong) output over the space $\boldsymbol{\mathcal{Y}}^\circ$ of configurations that are incompatible with the provided partial annotation, while the hinge loss searches throughout the entire space, encompassing both feasible and infeasible outputs.

### 3.4. Synthesis of Loss Functions

In fact, several other related loss functions, including ramp and max loss, have recently been proposed. They can be summarized in terms of a generic formulation:

$$\begin{aligned}
L^{\text{generic}}(x,y^*;\boldsymbol{w}) &= \Big| \max_{y\in\boldsymbol{\mathcal{Y}}^{\text{P}}}[f(x,y;\boldsymbol{w}) + \Delta(y^*,y)] \\
&\quad - \max_{y\in\boldsymbol{\mathcal{Y}}^{\text{R}}}[f(x,y;\boldsymbol{w})] \Big|_+ \quad (6)
\end{aligned}$$

Here, $\boldsymbol{\mathcal{Y}}^{\text{P}}$ is a "Penalty" space, since its members make a positive contribution to the loss. Accordingly, $\boldsymbol{\mathcal{Y}}^{\text{R}}$ denotes a "Reward" space because it contains the correct configuration and brings a negative contribution.

Table 2. Representation of related loss functions using the proposed generic formulation in Eq. 6.

| Loss | $\mathcal{Y}^P$ | $\mathcal{Y}^R$ | Appeared in Literature |
|---|---|---|---|
| hinge | $\mathcal{Y}$ | $\mathcal{Y}^*$ | (Yu & Joachims, 2009; Fernandes & Brefeld, 2011; Zhu et al., 2010; Vedaldi & Zisserman, 2009; Wang & Mori, 2010) |
| ramp | $\mathcal{Y}$ | $\mathcal{Y}$ | (Do et al., 2008; Girshick et al., 2011) |
| max | $\mathcal{Y}^\circ$ | $\mathcal{Y}$ | (Jie & Orabona, 2010) |
| bridge | $\mathcal{Y}^\circ$ | $\mathcal{Y}^*$ | This paper |

Table 2 provides a summary of related loss functions and shows how they fit into our generic formulation.

Those loss functions bear different properties. For example, while needed for max loss and bridge loss, the $|\cdot|_+$ operator can be dropped for ramp loss and hinge loss provided that $\Delta(\cdot,\cdot)$ is a positive function[2]. We refer the readers to (Zhang, 2004; McAllester & Keshet, 2011; McAllester et al., 2010) for more theoretical analysis on hinge/ramp loss.

### 3.5. Large Margin Learning Objective

With a clear definition of loss function in Eq. 6, we now set off to define the learning objective function. Take any loss function in Table 2 and insert it into Eq. 1. We obtain our learning objective as

$$\min_{\boldsymbol{w}} \quad \lambda\Omega(\boldsymbol{w}) + \underbrace{\frac{1}{N}\sum_n \max_{y\in\mathcal{Y}_n^P}[f(x_n,y_n;\boldsymbol{w}) + \Delta(y_n^*,y)]}_{P(\boldsymbol{w}),\text{ convex}}$$
$$- \underbrace{\frac{1}{N}\sum_n \max_{y\in\mathcal{Y}_n^R}[f(x_n,y;\boldsymbol{w})]}_{R(\boldsymbol{w}),\text{ convex}} \quad (7)$$

s.t.   each loss must be nonnegative.

Eq. 8 is a subtraction of two convex functions, namely $\lambda\Omega(\boldsymbol{w}) + P(\boldsymbol{w}) - R(\boldsymbol{w})$. Note that such structured learning problems are generally computationally expensive because the maximizations therein have to be solved at each iteration of updating $\boldsymbol{w}$ for each training sample. Next, we will present an efficient method to address this problem.

---

[2]For these losses, $\mathcal{Y}^R \subseteq \mathcal{Y}^P$, so the margin can never be negative.

## 4. Optimization with Bound Recycling

### 4.1. Convex-Concave Problem and CCCP

The difference of two convex functions forms a convex-concave optimization problem that can be solved by the CCCP procedure (Yuille & Rangarajan, 2003). Briefly, CCCP iterates between two steps:

**Step 1**: At iteration $t$, estimate a linear upper bound on the concave function $-R(\boldsymbol{w})$ using its subgradient at $\boldsymbol{w}_t$, viz. $\boldsymbol{v} = -\partial_{\boldsymbol{w}} R(\boldsymbol{w}_t)$. Then,

$$\langle \boldsymbol{v}_t, \boldsymbol{w} - \boldsymbol{w}_t \rangle - R(\boldsymbol{w}_t) \geq -R(\boldsymbol{w}), \forall \boldsymbol{w} \quad (8)$$

**Step 2**: Update the model by

$$\boldsymbol{w}_{t+1} = \arg\min_{\boldsymbol{w}} \tilde{J}(\boldsymbol{w}) := \lambda\Omega(\boldsymbol{w}) + P(\boldsymbol{w}) + \langle \boldsymbol{v}_t, \boldsymbol{w} \rangle. \quad (9)$$

The procedure is guaranteed to converge to a local minimum or saddle point (Yuille & Rangarajan, 2003).

(Yu & Joachims, 2009) used this strategy to optimize their structured SVM with latent variables, with a proximal bundle method (Kiwiel, 1990) for Step 2. (Girshick et al., 2011) and (Jie & Orabona, 2010) coined a similar procedure and applied stochastic gradient descent to speed up the training.

To construct the hyperplane for bounding $-R(\boldsymbol{w})$, one first solves $\tilde{y}_n = \arg\max_{y \in \mathcal{Y}_n^R}[f(x_n, y_n; \boldsymbol{w}_t)]$ for every $n$, and then computes $\boldsymbol{v}_t$ as

$$\boldsymbol{v}_t = \frac{1}{N}\sum_n \partial_{\boldsymbol{w}} f(x_n, \tilde{y}_n; \boldsymbol{w}_t) \quad (10)$$

### 4.2. Speeding Up CCCP with Bounds Recycling

Structured learning is computationally expensive due to the repetitive maximization problems one has to solve at every iteration to compute the subgradients (Tsochantaridis et al., 2006; Teo et al., 2010). This makes the above CCCP based optimization strategy particularly expensive because a complete structured learning has to be solved largely from scratch. We now introduce a novel method for speeding up CCCP when structured learning is required.

We first inspect the structure of the objective $\tilde{J}(\boldsymbol{w})$ in Eq. 9 and obtain the following key observations:

**Complexity**: $\tilde{J}(\boldsymbol{w})$ consists of three terms with different complexity: a regularizer $\lambda\Omega(\boldsymbol{w})$ (e.g., quadratic when using L2 regularization) and a linear term $\langle \boldsymbol{v}, \boldsymbol{w} \rangle$, both smooth and easy to solve, and a complicated, possibly non-smooth term $P(\boldsymbol{w})$.

**Consistency**: $\tilde{J}(\boldsymbol{w})$ changes at each CCCP iteration, due to the update of $\boldsymbol{v}$; however, the difficult function $P(\boldsymbol{w})$ remains the same.

Structured Learning from Partial Annotations

These two observations lead to two ideas for speedup.

Firstly, we construct a piecewise linear lower bound on the difficult $P(\boldsymbol{w})$ only, rather than on the entire objective $\tilde{J}(\boldsymbol{w})$ as in (Yu & Joachims, 2009). Since the $P(\boldsymbol{w})$ part of $\tilde{J}(\boldsymbol{w})$ remains the same, we can reuse these bounds across multiple CCCP iterations and avoid recomputing them from scratch. When some "good" linear approximation for $P(\boldsymbol{w})$ is provided at each iteration, solving $\tilde{J}(\boldsymbol{w})$ is easy because the other two terms are simple. We name this technique bounds recycling, since the bounds will be reused to compute the approximation gap between the original objective and its linear approximation.

Secondly, (Yuille & Rangarajan, 2003) showed that CCCP iteratively matches points on the two convex functions (i.e. $\lambda\Omega(\boldsymbol{w}) + P(\boldsymbol{w})$ and $R(\boldsymbol{w})$) which have the same subgradient, see Fig. 2 (left). Since we usually start with some $\boldsymbol{w}_0$ far from the optimum, it is not sensible to solve $\tilde{J}(\boldsymbol{w})$ to high precision at early iterations. Otherwise, many bounds need be computed to achieve this precision at some immature $\boldsymbol{w}$, which are mostly not reused at later iterations when precision really matters. Therefore, we propose to adaptively increase the precision of CCCP iteration until reaching the required precision. This procedure, named adaptive precision, is shown in Fig. 2 (right).

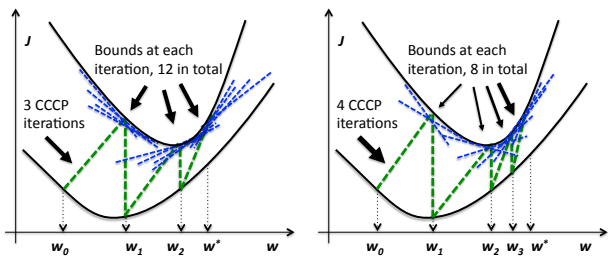

Figure 2. CCCP procedure: starting from $\boldsymbol{w}_0$, iteratively match points in the two curves which have the same subgradient, until convergence to the optimal $\boldsymbol{w}^*$. CCCP with fixed precision (left) requires fewer iterations, but more bounds than CCCP with adaptive precision (right). (best view in color)

### 4.3. Solving Model Update in the Dual

To construct a lower bound approximation for $P(\boldsymbol{w})$, we follow the bundle minimization method from (Teo et al., 2010). Briefly, at some $\boldsymbol{w}_k$, we compute the subgradient of $P(\boldsymbol{w})$ and the corresponding offset,

$$\boldsymbol{a} = \frac{1}{N}\sum_n \partial_{\boldsymbol{w}} f(x_n, \hat{y}; \boldsymbol{w}_k) \quad (11)$$

$$b = \frac{1}{N}\sum_n [f(x, \hat{y}; \boldsymbol{w}_k) + \Delta(y^*, \hat{y})] - \langle \boldsymbol{a}, \boldsymbol{w}_k\rangle \quad (12)$$

where $\hat{y} = \arg\max_{y\in\mathcal{Y}^P}[f(x, y; \boldsymbol{w}_k) + \Delta(y^*, y)]$ is the expensive augmented inference problem (Tsochantaridis et al., 2006). Now, this lower bound sitting at $\boldsymbol{w}_k$ can be expressed as $\langle \boldsymbol{a}, \boldsymbol{w}\rangle + b \leq P(\boldsymbol{w}), \forall \boldsymbol{w}$.

We store all subgradients $\boldsymbol{a}$ as column vectors in $\boldsymbol{A} = [\boldsymbol{a}_0, \boldsymbol{a}_1, \ldots]$ and the offsets $b$ in $\boldsymbol{b} = [b_0, b_1, \ldots]'$. Given $\boldsymbol{A}$ and $\boldsymbol{b}$, solving $\tilde{J}(\boldsymbol{w})$ in Eq. 9 becomes

$$\min_{\boldsymbol{w}} \ \lambda\Omega(\boldsymbol{w}) + \underbrace{\max_{(\boldsymbol{a},b)\in(\boldsymbol{A},\boldsymbol{b})}(\langle \boldsymbol{a}, \boldsymbol{w}\rangle + b)}_{\text{Linearly lower bounded } P(\boldsymbol{w})} + \langle \boldsymbol{v}, \boldsymbol{w}\rangle \quad (13)$$

Given regularizer $\Omega(\boldsymbol{w}) = \frac{1}{2}\|\boldsymbol{w}\|^2$, this problem can be easily solved in its dual form:

**Theorem 1.** *Given a list of lower bounds for some convex function expressed by subgradients $\boldsymbol{A} = [\boldsymbol{a}_0, \boldsymbol{a}_1, \ldots]$ and offsets $\boldsymbol{b} = [b_0, b_1, \ldots]'$, the dual form of the primal minimization problem in Eq. 13 is*

$$\max_{\boldsymbol{\alpha}} \ -\frac{1}{2\lambda}\boldsymbol{\alpha}'\boldsymbol{A}'\boldsymbol{A}\boldsymbol{\alpha} + \left(\boldsymbol{b}' - \frac{1}{\lambda}\boldsymbol{v}'\boldsymbol{A}\right)\boldsymbol{\alpha} \quad (14)$$
$$\text{s.t.} \quad \boldsymbol{\alpha}'\mathbf{1} = 1, \boldsymbol{\alpha} \geq \mathbf{0}.$$

*The primal variable $\boldsymbol{w}$ is connected to $\boldsymbol{\alpha}$ by*

$$\boldsymbol{w} = -\frac{1}{\lambda}(\boldsymbol{v} + \boldsymbol{A}\boldsymbol{\alpha}) \quad (15)$$

*Proof.* Very similar to Theorem 2 in (Teo et al., 2010). □

This dual form can be easily inserted into popular QP solvers such as CPLEX[3] and libqp[4].

### 4.4. Pseudocode and Implementation Details

Pseudocode of our optimization method is illustrated in Algorithm 1. We use $t$ to index CCCP iterations and $k$ to index lower bounds. Given rate $\rho \in (0,1)$, line 5 gradually increases the desired precision at each iteration until $\epsilon_{\min}$ (smaller means higher precision). Line 8 shows the accumulation of bounds that are reused every time at line 9. The approximation gap $\hat{\epsilon}$ is the margin between the original objective $\tilde{J}_t(\boldsymbol{w})$ and its lower bounded approximation, i.e. the minimum value of Eq. 13. We refer the readers to (Teo et al., 2010) for more details. Finally, the algorithm terminates when the decrease of the objective $\tilde{J}_t(\boldsymbol{w})$ between two consecutive CCCP iterations is smaller than some threshold $\eta$. Matlab code will be available to the public at http://xinghua-lou.org/research/.

It is important to point out that the nonnegativity constraint on each empirical loss must not be violated throughout the entire CCCP procedure. This is satisfied for hinge loss and ramp loss by their definition (McAllester & Keshet, 2011). For max loss and bridge loss, this can be achieved by ignoring samples that violate this constraint from the subgradient computation, as in usual SVM.

---
[3] http://www.ibm.com/software/
[4] http://cmp.felk.cvut.cz/~xfrancv/libqp/html/

**Algorithm 1** CCCP with Bounds Recycling
1: **Input:** $\{x_n, y_n^*\}$, $w_0$, $\eta$, $\{\epsilon, \epsilon_{\min}, \rho\}$
2: Initialize $t = 0, k = 0, A = \emptyset, b = \emptyset, w = w_0$
3: **repeat**
4:     Compute $v_t$ as in Eq. 10
5:     Set $\epsilon = \max(\epsilon \times \rho, \epsilon_{\min})$
6:     **repeat**
7:        Compute $a_k$ and $b_k$ as in Eq. 11 and Eq. 12
8:        Set $A = A \cup a_k$ and $b = b \cup b_k$
9:        Update $w$ using Eq. 13 with $A$, $b$ and $v_t$
10:       Compute approximation gap $\hat{\epsilon}$
11:       Set $k = k + 1$
12:     **until** $\hat{\epsilon} \leq \epsilon$
13:     Set $w_{t+1} = w$
14:     Set $t = t + 1$
15: **until** $\tilde{J}(w_{t-1}) - \tilde{J}(w_t) \leq \eta$
16: **Output:** $w$

## 5. Experiments

We evaluate our method on a very challenging real world problem: cell tracking. Robust tracking is of fundamental importance for, i.a., molecular, cell and developmental biology. Recently, (Lou & Hamprecht, 2011) proposed a structured learning for cell tracking which allows to *learn* the parameters of an energy function from manually annotated tracks, leading to significantly improved performance especially if the number of parameters becomes large. However, their learning strategy was based on classic structured learning, requiring *exhaustive* assignment annotations of pairs of frames. This is a tedious task at best, and becomes impossible for large scale problems.

### 5.1. Model, Data and Comparison Setup

(Lou & Hamprecht, 2011) formulate tracking by assignment as a constrained binary energy minimization problem. A foregoing detection step finds potential cells/targets in either of two consecutive frames. Based on these detections, a set $E$ of possible events (such as motion, division, etc.), described by features $\phi_{c,c'}^e$ is compiled. The indicator variables $y_{c,c'}^e$ state if an event is realized or not. Many events are mutually exclusive according to conservation laws: each detected cell must have a unique history and a unique fate. In summary, given the learned parameters $w$, a predicted tracking is obtained as the minimizer of

$$\min_y f(x, y; w) := \sum_{e \in E} \sum_{c \in C} \sum_{c' \in C'} \langle \phi_{c,c'}^e, w^e \rangle y_{c,c'}^e \quad (16)$$

$$\text{s.t. } \forall c' \in C', \sum_{e \in E} \sum_{c \in C} y_{c,c'}^e = 1 \text{ (conservation)} \quad (17)$$

$$\forall c \in C, \sum_{e \in E} \sum_{c' \in C'} y_{c,c'}^e = 1 \text{ (conservation)} \quad (18)$$

$$\forall e \in E, c \in C, c' \in C', y_{c,c'}^e \in \{0, 1\} \text{ (Booleanity)}$$

Here, $C$ and $C'$ are power sets of all detections from the respective frames to accommodate the description of events such as division, where one cell in the first frame can be assigned to two cells from the second frame.

To make the problem realistic and even harder, training[5] and test[6] data (both publicly available) from different (!) experiments and labs were used. The cardinality of the structured output (number of indicator variables) ranges from 400 at early stages to over 5000 at late stages, and the inference problem involves higher-order constraints up to order 50.

### 5.2. Comparison to Structured Perceptron and Full Annotation

To obtain a baseline, 20 randomly selected but fully annotated pairs of frames were used to train the model from (Lou & Hamprecht, 2011) using bundle minimization. Next, to obtain partial annotations, a variable fraction of all events was selected using stratified[7] random sampling. For better statistics, each experiment was repeated 10 times using different stratified random samples. To make all experiments comparable, the same precision (i.e., approximation gap, see Algorithm 1) was used for bundle minimization and the method proposed here. The structured perceptron with partial annotations was trained until the task loss became zero, or stopped improving, and no early stopping was used.

Fig. 3 shows a comparison of the average test loss. One surprising result is that the model learned from partial annotations as suggested here apparently can outperform the model learned from full annotation (Lou & Hamprecht, 2011) when only around 40% of all data is annotated. Our interpretation is that significantly less data may not be enough to optimally train the tracking model with its around 40 parameters; while significantly more data may lead to overfitting. Note that this phenomenon was also observed by (Fernandes & Brefeld, 2011). Secondly, the proposed method consistently outperforms the structured perceptron with partial annotation. We attribute this to the perceptron's lack of regularization, and resulting overfitting.

---

[5]http://www.cbi-tmhs.org/Dcelliq/files/051606_HeLaMCF10A_DMSO_1.rar
[6]http://www.mitocheck.org/cgi-bin/mtc?action=show_movie;query=24386
[7]Making sure that rare events such as division could become part of a partial annotation.

Fig. 4 shows a comparison of training times. Once the proportion of partial annotation exceeds 20%, our method requires roughly twice as much time as the bundle method for risk minimization that is working on full annotations only. Training the structured perceptron appears to be more expensive, but its runtimes have a lower variance.

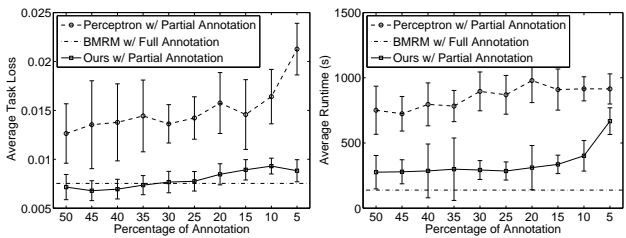

Figure 3. Comparison of average test loss.  Figure 4. Comparison of training time.

### 5.3. Comparison of Surrogate Losses

Table 3 shows a comparison of various loss functions w.r.t. prediction accuracy and runtime for the partially annotated data. We see that bridge (proposed here) and hinge loss yield very similar prediction performance, with somewhat faster runtime of the former.

Surprisingly, despite their very similar formulations, both ramp and max loss give much lower accuracy (around threefold higher test loss), but allow two- or threefold faster training. Recall that Table 2 shows the key difference between max/ramp loss and hinge/bridge loss: the former search through the entire space for the best configuration, while the latter only search within a subspace that is consistent with the partial annotation. Our result suggests that constraining the search to a feasible subspace that is compatible with the available annotations is crucial for the accuracy of the learned model.

To support this argument, we modify the ramp/max losses to make them aware of available partial annotations in their search for the highest-scoring configuration, i.e. when solving the second maximization in Eq. 6. This can be achieved by inserting a $-\Delta(y^*, y)$ into the second maximization (McAllester & Keshet, 2011), as

$$L^{\text{new}}(x, y^*; \boldsymbol{w}) = \left| \max_{y \in \mathcal{Y}^{\text{P}}} [f(x, y; \boldsymbol{w}) + \Delta(y^*, y)] - \max_{y \in \mathcal{Y}^{\text{R}}} [f(x, y; \boldsymbol{w}) - \Delta(y^*, y)] \right|_+$$

The performance of such modified ramp/max losses is shown in the last rows of Table 3. Their learning accuracy is significantly improved, and brought to the level of the hinge/bridge loss. Note that this modification has no effect on hinge/bridge loss.

Table 3. Comparison of loss functions.

| Loss Function | Test Task Loss (%) | Runtime (s) |
|---|---|---|
| Hinge | $0.85 \pm 0.055$ | $311.3 \pm 42.5$ |
| Ramp | $2.73 \pm 0.047$ | $165.4 \pm 67.9$ |
| Max | $2.71 \pm \mathbf{0.031}$ | $\mathbf{116.7} \pm 38.4$ |
| Bridge | $\mathbf{0.83} \pm 0.071$ | $254.1 \pm 28.8$ |
| Ramp - $\Delta(y^*, y)$ | $1.14 \pm 0.482$ | $257.5 \pm 48.6$ |
| Max - $\Delta(y^*, y)$ | $1.00 \pm 0.417$ | $191.5 \pm \mathbf{21.1}$ |

### 5.4. Comparison of Optimization Strategy

We compare our optimization strategy to the CCCP procedure from (Yu & Joachims, 2009) which does not use the bounds recycling and adaptive precision proposed here. In a lesion study, we also study the effect of omitting either bounds recycling or/and adaptive precision.

Fig. 5 shows the convergence of the objective function. All optimization methods converge to the same objective value. Using both bounds recycling and adaptive precision, we achieve a speed-up of a factor of 5 or so. Note that we implemented (Yu & Joachims, 2009)'s C-CCP procedure using the BMRM method (Teo et al., 2010) whose complexity $\mathcal{O}(\frac{1}{\epsilon})$ is actually better than that of the proximal bundle method used in the original paper, $\mathcal{O}(\frac{1}{\epsilon^3})$.

Fig. 6 shows the total number of bounds computed across the CCCP iterations. By using bounds recycling, our method only requires ca. 100 bounds until convergence, while (Yu & Joachims, 2009)'s approach computes almost 100 bounds at its first iteration.

## 6. Conclusions and Outlook

We conclude that structured learning from partial annotations is practically possible. With a proper choice of loss function and optimization strategy, the model learned from partial annotations has an accuracy that compares well with that obtained from exhaustive annotation.

Overall, we witness a fundamental tradeoff: in our experiments, successful learning from partial annotations is cheaper for the human by a factor of 2-4 (lower labeling effort), but more expensive for the computer by a similar factor. Given that we value human time more highly, and that labeling takes of the order of hours whereas computations are in the order of minutes, we believe that learning from partial annotations as proposed here and in (Fernandes & Brefeld, 2011), as well as implicitly in (Yu & Joachims, 2009), is fundamentally a sound idea that is worth pursuing.

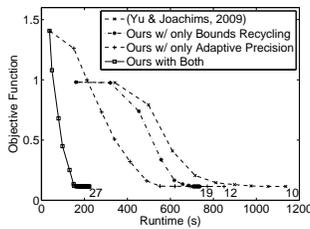
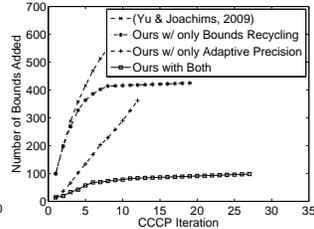

*Figure 5.* Decrease of the objective function.

*Figure 6.* Total Number of bounds before convergence.

In the future, we are interested in marrying our approach with active learning: that is, let the computer identify relevant partial structures which, if annotated, can reduce ambiguity in the current labeling, and help achieve steeper learning curves.